\documentclass[conference]{IEEEtran}
\IEEEoverridecommandlockouts
\usepackage{cite}
\usepackage{amsmath,amssymb,amsfonts}
\usepackage{algorithmic}
\usepackage{graphicx}
\usepackage{textcomp}
\usepackage{xcolor}
\usepackage{changes}   % remove "final" to show changes  % za hlchange komandu
\usepackage{svg}

\def\BibTeX{{\rm B\kern-.05em{\sc i\kern-.025em b}\kern-.08em
    T\kern-.1667em\lower.7ex\hbox{E}\kern-.125emX}}
\begin{document}

\title{ End-to-End O-RAN Testbed for Edge-AI-Enabled\\ 5G/6G Connected Industrial Robotics\\

}

\author{\IEEEauthorblockN{Sasa Talosi\IEEEauthorrefmark{2}, Vladimir Vincan\IEEEauthorrefmark{1}, Srdjan Sobot\IEEEauthorrefmark{2}, Goran Martic\IEEEauthorrefmark{2}, Vladimir Morosev\IEEEauthorrefmark{2}\\ Vukan Ninkovic\IEEEauthorrefmark{1}\IEEEauthorrefmark{2}, Dragisa Miskovic\IEEEauthorrefmark{1}, Dejan Vukobratovic\IEEEauthorrefmark{2} 
\vspace{1mm}
\IEEEauthorblockA{
\IEEEauthorblockA{\IEEEauthorrefmark{2}Faculty of Technical Sciences, University of Novi Sad, Serbia}
\IEEEauthorblockA{\IEEEauthorrefmark{1}The Institute for Artificial Intelligence Research and Development of Serbia, Serbia
}
}}}

\maketitle

\begin{abstract}
Connected robotics is one of the principal use cases driving the transition towards more intelligent and capable 6G mobile cellular networks. Replacing wired connections with highly reliable, high-throughput, and low-latency 5G/6G radio interfaces enables robotic system mobility and the offloading of compute-intensive artificial intelligence (AI) models for robotic perception and control to servers located at the network edge. The transition towards Edge AI as a Service (E-AIaaS) simplifies on-site maintenance of robotic systems and reduces operational costs in industrial environments, while supporting flexible AI model life-cycle management and seamless upgrades of robotic functionalities over time. In this paper, we present a 5G/6G O-RAN–based end-to-end testbed that integrates E-AIaaS for connected industrial robotic applications. The objective is to design and deploy a generic experimental platform based on open technologies and interfaces, demonstrated through an E-AIaaS-enabled autonomous welding scenario. Within this scenario, the testbed is used to investigate trade-offs among different data acquisition, edge processing, and real-time streaming approaches for robotic perception, while supporting emerging paradigms such as semantic and goal-oriented communications. 
\end{abstract}


\section{Introduction}

Connected robotics is one of the primary use cases for the ultra-reliable and low-latency communication (URLLC) service in 5G networks \cite{urllc}, and it continues to be considered a key driver for the evolution of URLLC towards hyper-reliable low-latency communications (HRLLC) in 6G \cite{hrllc}. Use case analyses, system requirements, and architectural concepts for 6G connected robotics have recently been studied \cite{6Grob1, 6Grob2}, including novel standardisation activities and initiatives \cite{6Gstd1, 6Gstd2}. The integration of collaborative multi-robot systems across private 5G/6G networks has been investigated in factory environments \cite{6Gmrs1} and in search-and-rescue missions \cite{6Gmrs2}. One of the most prominent use cases within the context of connected robotics is remote tele-surgery, which has been explored in several research studies \cite{6Gts1, 6Gts2}.

In contrast to research works investigating different 5G/6G connected robotic scenarios, comparatively fewer publications address connected robotic testbeds. The work in \cite{5Grt1} is an early example that explores connected robotics within a private 5G network testbed based on commercial gNB infrastructure. With the evolution of O-RAN standards \cite{orantut}, the focus has shifted towards O-RAN-compatible testbeds \cite{orantbed}. The work closest to ours is a recent demonstration of a remote tele-operation O-RAN testbed described in \cite{ORANtb1}. However, unlike tele-operation, our testbed focuses on exploiting semantic communication for industrial robotic applications. A similar focus on O-RAN and semantic communication, albeit with less emphasis on testbed infrastructure, is presented in \cite{ORANtb2}.

In this paper, we present an end-to-end O-RAN–compliant 5G/6G testbed designed to support edge-AI-enabled connected industrial robotics. The proposed platform leverages open interfaces and modular components to realize an Edge AI as a Service (E-AIaaS) paradigm, enabling flexible deployment, execution, and evolution of AI-driven robotic functionalities at the network edge. The testbed is demonstrated through an autonomous welding use case, where it is used to systematically investigate trade-offs among different data acquisition, edge processing, and real-time streaming strategies for robotic perception. In doing so, the platform provides a practical experimental framework for studying emerging concepts such as semantic and goal-oriented communications in realistic industrial robotic scenarios.

\section{Connected Robotics Use Case: Edge-AI Empowered Autonomous Welding}

The convergence of today’s robotics, 3D vision systems, and edge-cloud artificial intelligence is enabling a new generation of connected robotic workcells capable of perception-driven and adaptive operations. Unlike traditional industrial automation, where robots execute pre-programmed trajectories in isolated environments, connected robotics systems operate within dynamic production contexts, continuously exchanging data with cloud intelligence services and other machines.

\begin{figure}[!t]
\centering
\includegraphics[width=0.8\columnwidth]{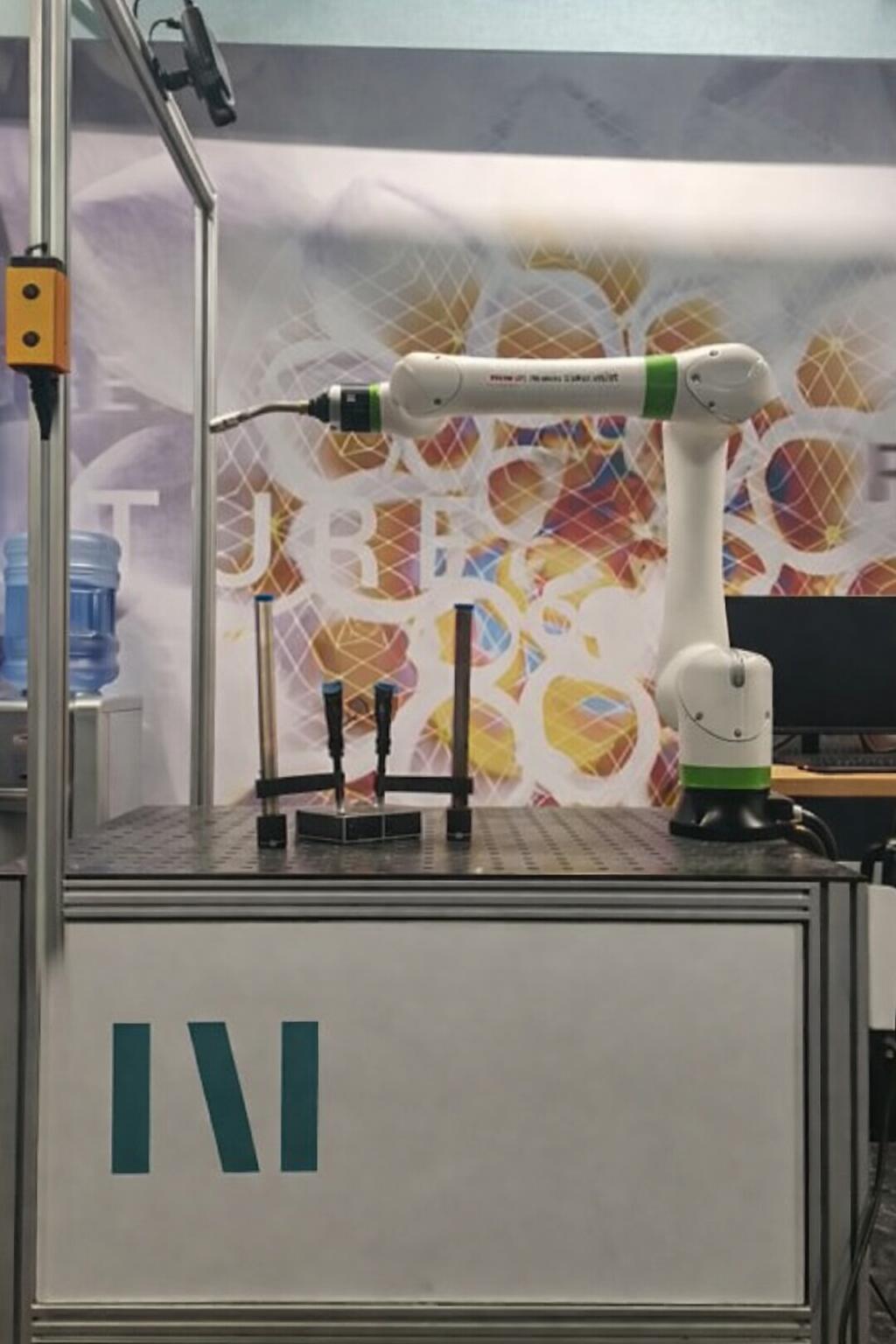}
\caption{Robotic workcell setup for AI-enabled autonomous welding.}
\label{Fig_TB}
\end{figure}

In this paper, we consider the AI-enhanced welding control system in which a collaborative robot (cobot) is tightly integrated with a high-resolution 3D vision system and networked computing infrastructure.  The connected robotic system we consider consists of the following components: 1) vision system: that captures 3D object representation, 2) robotic platform: robotic hand that executes robotic tasks, 3) workspace setup: the space where 3D vision is enabled and robotic tasks can be executed, 4) control server: executing local compute tasks, 5) 5G/6G network: providing necessary connectivity, and 6) remote server: executing remote edge AI tasks. An example of a robotic working cell (that includes components 1)--4) above) that we use in the testbed described in this paper (see details in Sec. IV) is illustrated in Fig. \ref{Fig_TB}.
%The functional diagram of connected robotic use case is illustrated in Fig. X.   

\subsection{Use Case Description}
The 3D sensor provides dense geometric and semantic information about the workspace and scene topology. AI models deployed at the edge process this multimodal data in real time to enable robust object detection, 6D pose estimation and scene understanding. The resulting perception outputs are transmitted to higher-level reasoning and orchestration layers, where task decisions, skill selection, and performance monitoring are coordinated across the connected system.

The perception pipeline starts with instance segmentation, isolating the welding workpiece from the surrounding scene. Because welding clamps are permanently present on the table, the system detects them using ArUco markers and removes their geometry from the point cloud. This prevents clamps from being misclassified as part of the workpiece and designates their regions as forbidden zones for collision-free motion planning.

The filtered point cloud is then processed by a 6D pose estimation module, which determines the object’s position and orientation in the camera coordinate frame. In the next step, the system determines the object’s 6D pose and geometry for downstream trajectory generation.
After the perception pipeline completes processing of the scene snapshot, the system transforms the object pose and geometry into the robot coordinate system. That is a starting point for generating robot movement along the optimal path.

\subsection{Use Case Requirements}

The communication requirements of the autonomous welding scenario are dictated by high-resolution 3D perception and closed-loop trajectory adaptation. Unlike conventional industrial automation, where motion paths are pre-programmed, the considered system continuously captures the scene, estimates object pose, and refines welding trajectories based on perception feedback. Consequently, throughput, latency, and reliability constraints must be jointly satisfied, as detailed in Table~\ref{summary_table} and the text below.

\begin{table}[t]
\centering
\caption{Use Case Requirements for Autonomous Welding}
\begin{tabular}{l c}
\hline
Parameter & Value / Range \\
\hline
Raw frame size & 30 MB \\
Frame rate & 2–10 FPS (mode dependent) \\
Raw uplink rate (5 FPS) & $\approx$ 1.2 Gbps \\
Raw uplink rate (10 FPS) & $\approx$ 2.4 Gbps \\
Compressed uplink (5 FPS) & 120–240 Mbps \\
Downlink rate & $<$ 5 Mbps \\
Traffic asymmetry & 95–99\% uplink \\
3D acquisition time & 200 ms – 700 ms \\
Comm. + edge latency & $<$ 100 ms \\
Emergency latency & $<$ 10 ms \\
Reliability & $\geq$ 99.999\% \\
\hline
\end{tabular}
\label{summary_table}
\end{table}

\subsubsection{3D Data Generation and Frame Rate}

The 3D sensing system produces dense point clouds containing approximately 5 million points per frame, corresponding to a spatial resolution of 2448 × 2048. Each point includes XYZ coordinates (millimeter precision), RGB color values (8-bit), and signal-to-noise ratio (SNR) information, resulting in approximately 30 MB of raw data per frame.
Industrial high-resolution 3D reconstruction systems based on structured-light or active stereo sensing typically operate in the range of 2–10 frames per second (FPS), depending on exposure configuration and reconstruction complexity \cite{p2, ros}. In perception-driven robotic welding, scene updates are not required at video rate, since the workpiece geometry remains quasi-static during each welding segment. Prior studies report perception update rates between 3–10 Hz as sufficient for closed-loop trajectory adaptation and seamless tracking \cite{p2, liu_2026}. A nominal operating point of 5 FPS therefore represents a realistic industrial configuration, while 10 FPS reflects an upper-bound stress configuration for network capacity evaluation.

\subsubsection{Traffic Characteristics}

The traffic asymmetry of the considered use case originates from the intrinsic imbalance between high-volume perception data generation and comparatively low-volume control information exchange. The 3D sensing system produces dense geometric representations of the workspace, while the edge-AI platform returns compact outputs such as 6D pose estimates, trajectory parameters, and supervisory commands. As a result, the communication load is inherently dominated by uplink transmission from the control server to the edge.

\begin{figure*}[!t]
\centering
\includegraphics[width=0.95\textwidth]{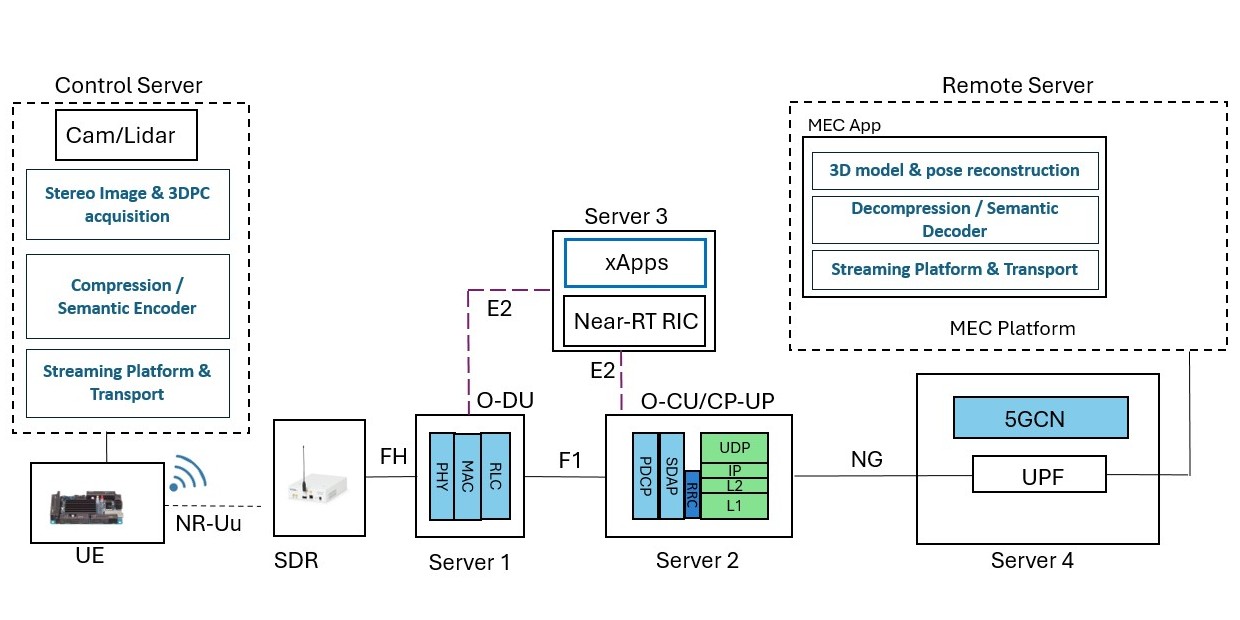}
\caption{End-to-end Connected Robotics O-RAN-based Testbed Architecture.}
\label{Fig_1}
\end{figure*}

Given a raw frame size of approximately 30 MB, a 5 FPS stream generates 150 MB/s ($\approx$1.2 Gbps), while 10 FPS requires approximately 2.4 Gbps. These values represent a worst-case upper bound corresponding to uncompressed, full-resolution streaming and serve as a reference point for radio capacity benchmarking. When geometric compression is applied (e.g., MPEG Point Cloud Compression (PCC)), a conservative compression ratio of 5:1–10:1 reduces the per-frame size to 3–6 MB~\cite{mpeg_2017}. At 5 FPS, the corresponding uplink data rate ranges between 120–240 Mbps. Further reduction is achievable by transmitting only task-relevant representations instead of full geometry, significantly lowering the uplink demand. The corresponding processing and encoding approaches are discussed in Section IV.

In contrast, downlink traffic consists of refined 6D pose estimates, welding trajectory parameters, motion waypoints, and supervisory control commands. Even in worst-case conditions, downlink data rates remain below 5 Mbps, typically below 1 Mbps. As a result, 95–99\% of the total traffic volume is concentrated in the uplink direction.

\subsubsection{Latency and Reliability Requirements}

Closed-loop welding adaptation requires that perception updates are reflected in robot motion within bounded temporal limits. In structured-light 3D sensing systems, the dominant component of the perception cycle is sensor acquisition and reconstruction.
Industrial structured-light cameras typically exhibit capture times ranging from approximately 200 ms in fast acquisition
modes to 700 ms in high-fidelity HDR configurations, depending on exposure settings and reconstruction accuracy~\cite{zivid_support}. This acquisition stage therefore determines the achievable perception update rate
and represents the primary contributor to overall system latency. Reduction of this latency is possible if visual system is not used in structured light mode and produces conventional RGB image/video data. 

Following frame acquisition, the communication and edge-processing pipeline must operate within tight timing constraints to prevent
additional delay accumulation. A representative latency allocation assumes uplink transmission below 20 ms, edge processing between
30–80 ms, and downlink response below 10 ms. Consequently, communication and computation contribute less than 100 ms
additional delay beyond the sensing cycle, ensuring that communication and inference do not dominate the perception cycle latency~\cite{3GPP_report}. For safety-critical operations such as emergency stop commands, latency must remain below 10 ms, consistent with ultra-reliable low-latency communication (URLLC) requirements for industrial motion control~\cite{popovski_2018}. The required end-to-end reliability is at least 99.999\%, corresponding to packet error rates below $10^{-5}$
during active welding.

% - brojevi : veličine slika, point cloudova, mapa...
% u pipelinu segmentacija i estimacija poze. Kakvi su zahtevi tu? Kakve resurse traži model. - processing time očekivan/ videti da chatgpt izgeneriše
% Da li se on može postaviti na edge-u ili može ići na cloud? Utvrditi.

% bandwidth
% latency
% asimetričan uplink/downlink saobraćaj
% uplink point cloud
% downlink npr poslati robotu komandu za poziciju

% navesti korake u pipeline-u i videti u stvarnom industrijskom okruženju koliko taj proces treba da traje.

% TODO: KOPIRANO IZ https://docs.google.com/document/d/194dOIxYets7Ias2cLAPOBtn7nzU8p8wb/edit
%The generated point cloud consists of 5 million points. Since there is a 1:1 correlation between pixels and points, it is possible to obtain XYZ (mm), RGB (8-bit), and SNR for every pixel, where SNR is the Signal-to-Noise Ratio. Internally on the GPU the 3D coordinates, color values, and SNR values are stored as separate 2D arrays of size 2448 x 2048 (around 30 MB per point cloud). Fig.~\ref{fig:pcd_representation} shows how point cloud and other data are represented.
% !TODO: KOPIRANO IZ https://docs.google.com/document/d/194dOIxYets7Ias2cLAPOBtn7nzU8p8wb/edit

%\begin{figure}[!ht]
%    \centering
%    \includegraphics[width=1.0\linewidth]{pcd_representation.png}
%    \caption{Point Cloud Representation}
%    \label{fig:pcd_representation}
%\end{figure}

\section{Testbed Architecture}

\subsection{3GPP/O-RAN 5G Network Testbed Architecture}

The end-to-end O-RAN-based testbed architecture is illustrated in Fig. \ref{Fig_1}. It consists of the part that hosts the 3GPP/O-RAN 5G network deployed on four telco-grade servers (Servers 1-4) and two application servers (Control and Remote) associated with the robotic use case.

\begin{figure*}[!t]
\centering
\includegraphics[width=0.95\textwidth]{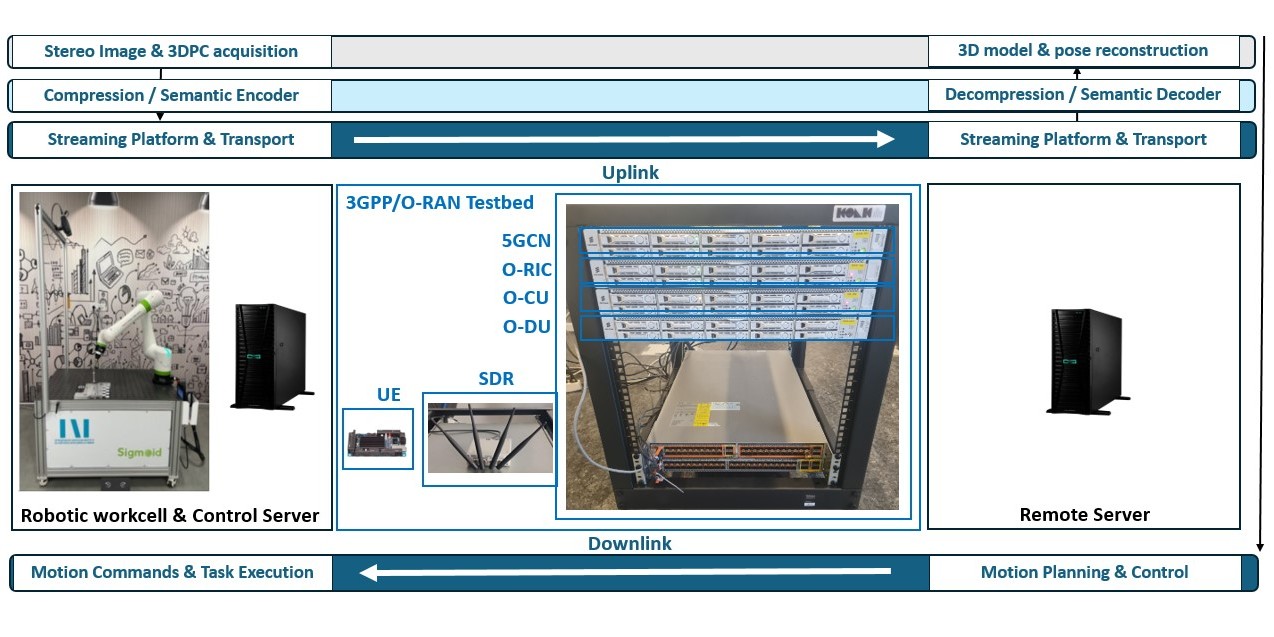}
\caption{End-to-end Connected Robotics O-RAN-based Testbed Components and Deployment.}
\label{Fig_2}
\end{figure*}

The 5G testbed follows the 3GPP/O-RAN architecture with a disaggregated base station (gNB). The base station is divided into a distributed unit (O-DU) and centralized unit (O-CU) interconnected via F1 interface. O-CU runs separate entities for user plane (O-CU-UP) and control plane (O-CU-CP) traffic. O-DU interfaces to standard software defined radios (SDR) as radio units via fronthaul (FH) - functional split 8 interface. O-CU interfaces towards the 5G core network (5GCN) via backhaul NG-C/NG-U interface for user and control plane traffic. According to O-RAN architecture, O-CU and O-DU are managed by Radio Interface Controllers (RIC). Herein, we consider only near-real-time (Near-RT) RIC that connects to O-DU and O-CU via E2 interface. Near-RT RIC hosts custom-designed extended applications (xApps) that could be designed to efficiently manage downlink/uplink traffic of a given use case in order to satisfy its use case requirements. User plane data going to/from the base station passes through breakout user plane function (UPF) and is forwarded to the Remote server located at the edge. The rest of the 5GCN is a standard service-based architecture comprising a large number of relevant core network functions (NFs). More details about specific deployment of 3GPP/O-RAN 5G network testbed is provided in Subsection IV-A below.

\subsection{Edge-AI Empowered Robotic Testbed Architecture} 

Robotic part of the testbed consists of the following elements: 1) industrial robotic hand, 2) sensing system (camera, lidar) for 3D object capture, 3) Control server, and 4) Remote Server (Fig. \ref{Fig_1}). The on-premise control server interfaces both the robotic hand and the sensing system to manage data coming to/from these elements. To transmit data via 5G testbed to the remote server, the control server interfaces 5G user equipment (UE) or modem. In the current testbed version, our focus is on the processing and transmission of data captured by the sensing system due to its significant compute and communication requirements. In this context, sensing system periodically capture raw image and 3D point cloud (PC) frames that are delivered with a given frame rate to the control server for further processing. The following data processing block performs classical image/3DPC compression. Alternatively, for research purposes, we consider the possibility of semantic data encoding tailored to a given edge AI task to be executed at the remote server. The resulting compressed/semantically encoded data is delivered to real-time streaming platform that handles data packetisation and transport across the 5G network testbed. 

At the edge side, the data is delivered to a Mobile Edge Computing (MEC) platform at the remote server where it is first processed by the real-time streaming/transport platform. The compressed/semantically encoded packet data stream is then decompressed/semantically decoded. Finally, image/3DPC data or their relevant semantic representation is delivered to the edge AI model that deals with 3D model and pose reconstruction of the object placed on the robotic platform workspace. In the next subsection, we provide details on the deployment of each architectural element in our testbed.   

\section{Testbed Deployment}

\subsection{3GPP/O-RAN 5G Network Testbed Deployment} 

For repeatable end-to-end testing, our 3GPP/O-RAN 5G network testbed is configured as a small, portable rack that combines radio/UE, switching, and computing equipment (Fig.~\ref{Fig_2}). With a single top-of-rack switch and four commodity servers (ran1-ran4), the rack allows for clean separation of fronthaul, backhaul, and RIC traffic via VLANs while maintaining local and controllable interconnects.

\subsubsection{Infrastructure for Computing and Switching}

Four Quanta D51B-1U servers, each with two Intel Xeon E5-2600 v3 processors, 32 GB of RAM, and a 500 GB hard drive, are used in the testbed. 10G SFP+ DAC (Direct Attach Copper) Twinax cables are used to connect each server to the Cisco N5K-C56128P switch. An external laptop that can SSH into every node for control and monitoring of the entire setup.

\subsubsection{VLAN segmentation and network topology}

\begin{figure}
    \centering
    \includegraphics[width=0.55\linewidth]{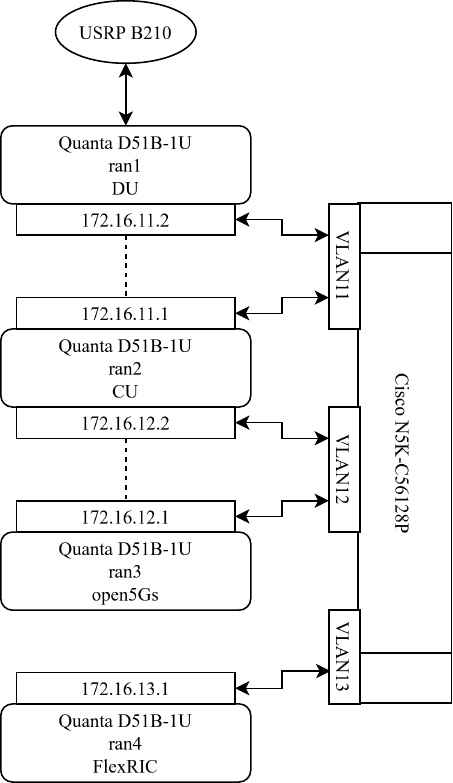}
    \caption{O-RAN Testbed Network Diagram}
    \label{fig:network_diagram}
\end{figure}

Three VLANs on the same switch: VLAN11 (fronthaul), VLAN12 (backhaul), and VLAN13 (RIC segment for ran4) provide logical separation. Cross-VLAN accessibility is possible because all VLANs terminate on the same physical switch, making end-to-end tracing and centralized supervision easier during experiments. Node addresses and connectivity are configured as follows (Fig.~\ref{fig:network_diagram}):
\begin{itemize}
    \item \textbf{ran1:} ens3f0 has IP 172.16.11.2 on VLAN11
    \item \textbf{ran2:} dual-homed with ens3f0 on VLAN11 with IP 172.16.11.1, and ens3f1 on VLAN12 with IP 172.16.12.2
    \item \textbf{ran3:} ens3f0 on VLAN12 with IP 172.16.12.1
    \item \textbf{ran4:} ens3f0 on VLAN13 with IP 172.16.13.1
\end{itemize}

%\begin{figure}[!ht]
%    \centering
%    \includesvg[width=0.55\linewidth]{network_diagram.svg}
%    \caption{O-RAN Testbed Network Diagram}
%    \label{fig:network_diagram}
%\end{figure}

\subsubsection{Radio and UE Setup}

On the radio side, a National Instruments USRP B210 with 4 antennas is directly connected to ran1 via USB 3.0. We use two commercial 5G UE kits (Quectel RMU500-EK) based on the RM500Q-GL modem. These kits offer reliable, consistent endpoints for connectivity and performance assessment. To support experiments at 100 MHz channel bandwidth and enable evaluation under higher-capacity radio configurations, we are concurrently expanding the platform with a USRP X300.

%\begin{figure}[!ht]
%    \centering
%    \includegraphics[width=0.85\linewidth]{RMU500-EVB-Kit.png}
%    \caption{Quectel RMU500-EK}
%    \label{fig:ue}
%\end{figure}

\subsubsection{Functional Split and Software Placement}

The RAN and other essential components are spread across dedicated servers in the testbed's Split 8 architecture. In particular, ran1 hosts the srsRAN DU~\cite{srsran_project} set up for TDD operation at 3825 MHz, 2×2 MIMO, and 20 MHz channel bandwidth; ran2 hosts the matching srsRAN CU~\cite{srsran_project}; and ran3 hosts the Open5GS-based 5G core~\cite{open5gs}. FlexRIC is deployed on ran4 and instantiated on VLAN13 to maintain a logical separation between RIC experimentation and the fronthaul/backhaul planes.

\subsection{Edge-AI Empowered Robotic Testbed Deployment} 

The robotics system part of the testbed is built upon the following specific hardware components (see Fig. \ref{Fig_TB}):

\subsubsection{Vision System} A Zivid 2+ MR130 3D camera, mounted in a fixed overhead position above the workstation, as shown in Fig.~\ref{Fig_TB}. Data acquisition procedure provides access to the raw RGB image, SNR image, depth map, and 3D point cloud.

\subsubsection{Robotic Platform} A FANUC CRX-10iA collaborative robot, selected for its safety features, adequate reach, and payload capabilities required for welding operations. The robot is controlled using a ROS 2–based setup integrated with the MoveIt motion planning framework. A customized MoveIt Task Constructor (MTC) pipeline is used to support welding-oriented motion generation, enabling the execution of trajectories derived from the AI perception module.

\subsubsection{Control and Remote Server} All perception inference, data processing, and trajectory generation run on two workstations used for model training and testing. The workstations are equipped with an NVIDIA RTX 4070 GPU, enabling efficient handling of high-resolution 3D data and AI inference tasks.

\subsubsection{Workspace Setup} The system operates on a fixed work table, with objects placed freely anywhere within the camera’s field of view (see Fig. \ref{Fig_TB}). This flexibility demonstrates the robustness of the AI-based detection approach.

\subsubsection{Compression/Semantic Encoding Layer}
The Compression/Semantic Encoding layer is introduced as an architectural enabler for flexible data representation within the proposed testbed. In the baseline configuration, raw RGB images and dense 3D point clouds are transmitted, representing a reference point with maximal bandwidth usage and minimal preprocessing overhead. The architecture also supports conventional geometric compression (e.g., MPEG PCC), which reduces uplink data rates at the cost of additional encoding complexity and delay at the control server. Furthermore, the design foresees task-oriented semantic encoding, where only information relevant to the downstream AI inference task, such as segmented regions or feature descriptors, is transmitted instead of full geometry. While this approach can substantially decrease communication load, it increases local computation and encoding latency, thereby affecting the overall perception-to-action delay budget. By enabling raw transmission, classical compression, and semantic encoding within a unified framework, the proposed architecture provides an extensible platform for studying compute–communication–latency trade-offs in 5G/6G connected robotic systems.

\subsubsection{Real-Time Streaming Layer} Compressed or semantically-encoded media/point-cloud streams should be efficiently streamed to remote edge AI server via 5G/6G network. Our architecture design and testbed experimentation assumes integration of compressed/semantic media with state-of-the-art real-time streaming platforms. Our current focus is on extension of Rust-based open-source selective forwarding unit platforms with WebRTC native client endpoints to support compression or semantic AI-based encoder data. 

\section{Sample Results and Future Work}

\begin{figure}
\centering
\includegraphics[width=\linewidth]{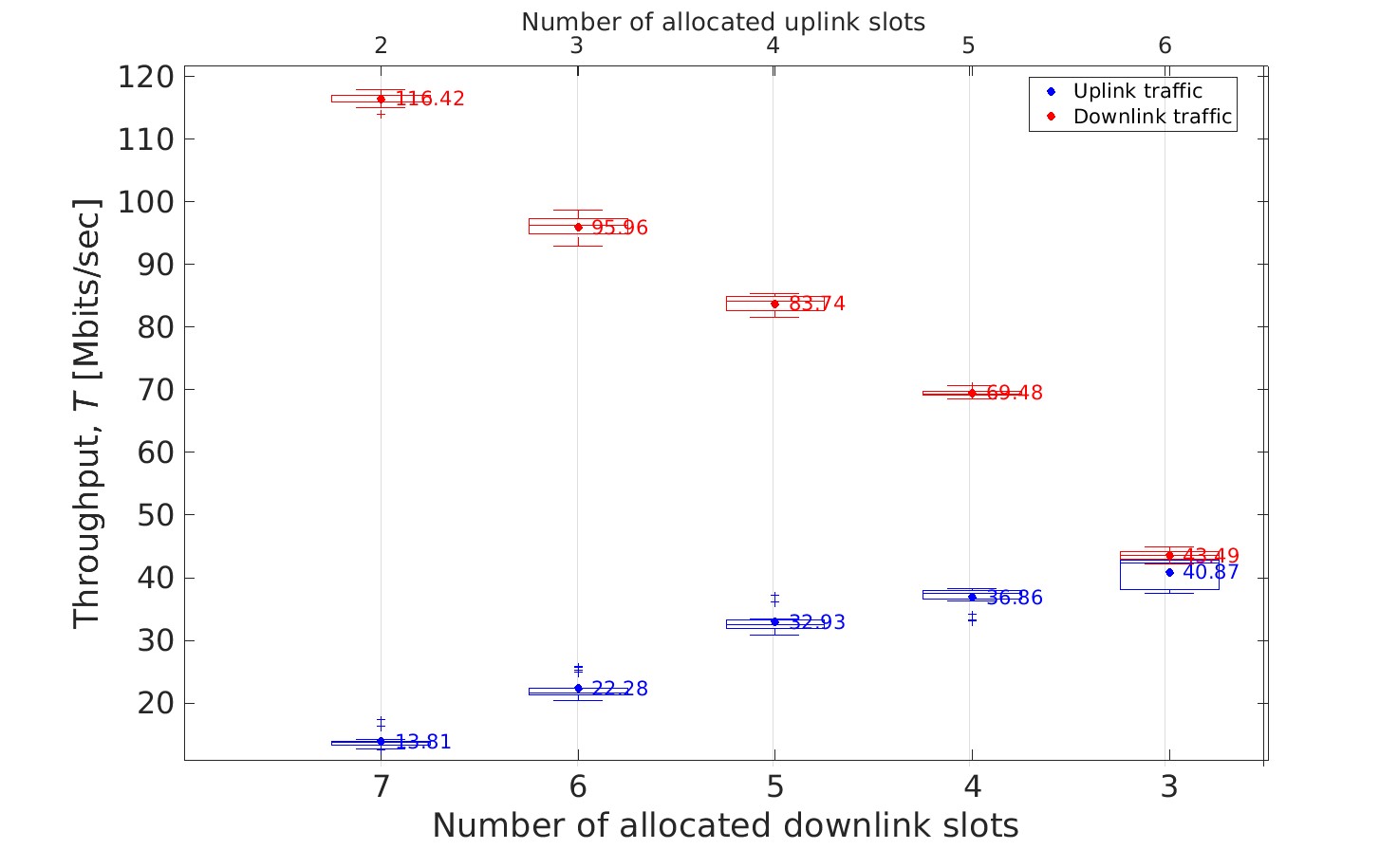}
\caption{Sample performance results for DL/UL throughput under different TDD slot configurations.}
\label{Fig_4}
\end{figure}

Although extensive testing and evaluation of complete end-to-end autonomous welding enabled by E-AIaaS will be provided in our future work, here we provide a sample of initial testing results. Fig. \ref{Fig_4} illustrates the downlink (DL) and uplink (UL) throughput with a varying balance between the DL/UL slot allocation in a TDD frame of 10 slots (where one slot is left unassigned). The throughput ranges between 43-116 Mbps for DL (2x2 MIMO) and 13-40 Mbps for UL (single-antenna setup) data for 20MHz PHY configuration performing under maximum channel quality (CQI being consistently reported at maximum level). As we migrate to the USRP X300 and 100 MHz channels, enabling 2x2 or 4x4 MIMO in the UL, we expect 5-20-fold increase in data rates that would provide adequate UL rate to support for a variety of compressed/semantically encoded media streams (as detailed in Sec. II.B). Our initial latency tests measured between the UE and the UPF demonstrate characteristic periodic latency variation between 16-32 ms due to NR uplink scheduling implementation in srsRAN. Further work and optimisation is required to properly minimise and adequately control both DL and UL latency. Principal directions of extenstion of the presented platform is: 1) migration to 100 MHz channels and 4x4 MIMO for throughput enhancement, 2) Integration of media compression/semantic encoding and WebRTC-based streaming for optimal cross-layer performance, 3) development of use-case optimal resource allocation xApps/rApps, 4) optimized media capture and acquisition, 5) testing and evaluation of various key performance/value indicators (KPI/KVIs) relevant for the use case, and 6) extensive investigation of compression/semantic encoding layer solutions and its integration with streaming layer for optimal end-to-end task performance. 

\end{document}